# Multi-modal user interface control detection using cross-attention

Milad Moradi[*]

AI Research Lab, Tricentis, Vienna, Austria

m.moradi-vastegani@tricentis.com

Ke Yan

AI Research Lab, Tricentis, Sydney, Australia

k.yan@tricentis.com

David Colwell

AI Research Lab, Tricentis, Sydney, Australia

d.colwell@tricentis.com

Matthias Samwald

Institute of Artificial Intelligence, Center for Medical Statistics, Informatics, and Intelligent Systems, Medical University of Vienna, Vienna, Austria

matthias.samwald@meduniwien.ac.at

Rhona Asgari

AI Research Lab, Tricentis, Vienna, Austria

r.asgari@tricentis.com

---

[*] Corresponding author. **Postal address:** Tricentis GmbH, Leonard-Bernstein-Straße 10, 1220 Vienna, Austria.

# Abstract

Detecting user interface (UI) controls from software screenshots is a critical task for automated testing, accessibility, and software analytics, yet it remains challenging due to visual ambiguities, design variability, and the lack of contextual cues in pixel-only approaches. In this paper, we introduce a novel multi-modal extension of YOLOv5 that integrates GPT-generated textual descriptions of UI images into the detection pipeline through cross-attention modules. By aligning visual features with semantic information derived from text embeddings, our model enables more robust and context-aware UI control detection. We evaluate the proposed framework on a large dataset of over 16,000 annotated UI screenshots spanning 23 control classes. Extensive experiments compare three fusion strategies, i.e. element-wise addition, weighted sum, and convolutional fusion, demonstrating consistent improvements over the baseline YOLOv5 model. Among these, convolutional fusion achieved the strongest performance, with significant gains in detecting semantically complex or visually ambiguous classes. These results establish that combining visual and textual modalities can substantially enhance UI element detection, particularly in edge cases where visual information alone is insufficient. Our findings open promising opportunities for more reliable and intelligent tools in software testing, accessibility support, and UI analytics, setting the stage for future research on efficient, robust, and generalizable multi-modal detection systems.

# 1. Introduction

User Interface (UI) control detection is critical for applications such as automated testing, accessibility, and software analytics [1]. In automated testing, reliably identifying UI elements like buttons and input fields ensures correct functionality across platforms [2]. For accessibility, detecting UI controls helps assistive technologies, such as screen readers, describe the interface to users with disabilities [3]. In software analytics, understanding user interactions with UI elements offers insights into usage patterns and design improvements [4]. However, detecting UI elements is challenging due to the high variability in UI designs, pixel-level similarity between elements, and the lack of large labeled datasets for training [2].

Many current UI detection systems rely primarily on pixel-based analysis, which can lead to ambiguities when UI controls share similar shapes, colors, or sizes [5]. For example, buttons, sliders, and input fields often appear visually similar, especially when they are small or have minimal distinguishing features, making it difficult for traditional models to accurately differentiate between them. This issue is compounded when dealing with highly cluttered or complex user interfaces where background elements may interfere with the identification of UI controls. As a result, purely visual approaches are prone to errors and inconsistencies, which affects their robustness across diverse and dynamically evolving UI designs [6].

Furthermore, traditional object detection methods struggle when visual cues alone are insufficient or when the context of an element is unclear. For example, in cases where the layout of UI elements changes based on user interaction or platform-specific design choices, relying solely on pixel-based features may not provide enough information for accurate detection [2]. In such situations, the lack of context can significantly hinder detection performance. This highlights the need for additional modalities, such as textual descriptions, to help disambiguate visual elements and provide contextual understanding [7]. Integrating textual data, such as descriptions of UI controls, can offer important contextual information that complements visual cues, improving detection accuracy and robustness [5, 8].

In recent years, GPT-based language models [9, 10] have shown remarkable capabilities in generating high-level or descriptive captions for various types of images, including software screenshots [11-13]. These models can produce detailed textual descriptions, such as "the screenshot contains a button labeled Submit" or "a text input field with a placeholder 'Enter your name'," that convey important information about the objects within the image. These captions go beyond simple pixel-based recognition by providing a higher-level semantic understanding of the content, which can be particularly useful in complex UI designs. By generating clear, concise descriptions of UI elements, GPT models can provide context that helps disambiguate visually similar objects, allowing for more accurate detection of UI controls [14].

Moreover, the semantic information embedded in the textual descriptions can be effectively captured using GPT embeddings [15], which map the descriptions into high-dimensional vector spaces. These embeddings can represent the contextual and relational aspects of UI elements, offering supplementary information that purely visual features might miss [16]. Combining these textual embeddings with visual features extracted from images enables the model to leverage both the visual cues and the contextual insights embedded in the text. This multimodal approach can improve detection accuracy, especially in challenging edge cases where visual features alone may be insufficient, such as when UI controls have minimal distinguishing visual differences or when the design changes dynamically. The fusion of text and image data can enable more robust performance, enhancing the model's ability to handle diverse and complex UI layouts.

Object detection models like YOLO (You Only Look Once) [17] have shown promise in detecting elements from graphical user interface images. YOLO's real-time detection and ability to identify multiple objects within an image make it well-suited for UI element detection [18, 19]. By leveraging the power of deep learning, YOLO can generalize across diverse UI designs and detect various types of UI elements. The combination of visual inputs with textual descriptions, such as those from GPT-based language models, can further enhance YOLO detection accuracy by providing additional context, improving robustness to design variability [8, 20].

In this work, we extend the YOLOv5 detection architecture into a multimodal framework that integrates GPT-generated textual descriptions of UI screenshots. The base YOLOv5 model consists of three main components: the backbone (responsible for hierarchical visual feature extraction), the neck (which aggregates multi-scale features), and the head (which outputs bounding boxes and class predictions). To incorporate textual information, we introduce cross-attention modules after the C3 blocks in the neck. These modules align the visual features from the backbone with semantic features from text embeddings derived from textual descriptions. The cross-attention mechanism allows the detector to prioritize visual features that correspond to semantically meaningful textual cues, effectively enabling the language modality to guide the visual detection process. The integration of text is expected to improve detection accuracy, especially in cases where visual features alone may be ambiguous, such as distinguishing between similarly appearing UI controls or handling edge cases that lack sufficient visual distinction.

To evaluate our proposed multi-modal YOLO framework, we conducted extensive experiments on a large UI control detection dataset comprising over 16,000 annotated screenshots with 23 distinct UI control classes. As a baseline, we trained a standard YOLOv5 model using only visual features. While this model performed strongly on frequently occurring and visually distinctive controls such as *Button*, *Image*, and *Text*, it struggled with semantically complex or visually ambiguous classes. These limitations highlighted the need for additional contextual information beyond pixel-level cues.

We then conducted various experiments on our multi-modal detection model, with three different fusion methods, i.e. element-wise addition, weighted sum, and convolutional fusion, for merging image and text features within cross-attention modules. Incorporating GPT-generated textual descriptions into the detection pipeline consistently improved performance across all fusion strategies. The element-wise addition approach provided immediate gains, particularly for underperforming classes, while weighted sum fusion delivered a balanced improvement across most metrics. The best results were obtained with convolutional fusion, representing a clear advancement over both the baseline and simpler fusion methods. Notably, this variant significantly improved detection of challenging classes such as *Horizontal_Axis*, *Vertical_Axis*, *Graph*, *Chart*, *Image*, and *Checkbox_Checked*, where purely visual cues had previously been insufficient. These results confirm that integrating textual semantics with visual features enables more robust and accurate UI element detection, particularly in cases of ambiguity or design variability.

The key contributions of this paper are as follows:

- **Introducing a novel multi-modal UI detection model:** We extend the YOLOv5 framework with cross-attention modules that effectively integrate image features with GPT-generated textual descriptions, enabling more robust and context-aware detection of UI controls from software screenshots.
- **Providing a comprehensive empirical evaluation:** We perform extensive experiments on a large-scale dataset of 16,000+ annotated UI screenshots spanning 23 control classes, demonstrating the practical value of our approach across a wide range of interface elements.
- **Analyzing fusion strategies for multi-modal learning:** We systematically investigate three alternative methods for merging image and text features, i.e. element-wise addition, weighted sum, and convolutional fusion, highlighting how the complexity of the fusion mechanism influences detection performance.
- **Achieving state-of-the-art performance in UI control detection:** We show that our multi-modal model consistently outperforms the baseline YOLOv5 across all key evaluation metrics, with particularly significant improvements on challenging and semantically ambiguous classes, underscoring the effectiveness of combining visual and textual modalities.

Together, these contributions establish a new direction for UI control detection by demonstrating how multi-modal integration can overcome the limitations of purely vision-based approaches and set a foundation for future advances in software testing, accessibility, and interface analysis.

The remainder of this paper is structured as follows. Section 2 reviews related work on UI element detection, object detection architectures, and multimodal fusion. Section 3 introduces our multimodal YOLO architecture, detailing the backbone, neck, and cross-attention integration. Section 4 describes the dataset, training setup, and evaluation methodology. The experimental results, ablation studies, and performance analysis are also presented and discussed in Section 4. Finally, Section 5 concludes with key findings and outlines directions for future work.

## 2. Related work

Detecting UI elements directly from screenshots has been a long-standing challenge due to the diversity of layouts and the visual similarity of many controls. Early work such as REMAUI [21] used heuristic image processing and OCR to identify UI components, but these methods lacked robustness across platforms and designs. With the rise of deep learning, object detection architectures have been increasingly applied to UI element recognition [19, 22]. Chen et al. [1] compared traditional approaches with deep object detectors such as Faster R-CNN, YOLO, and CenterNet, showing that modern detectors significantly outperform earlier methods. Subsequent works have adopted YOLO-based detectors for UI analysis, given their real-time performance and high accuracy [7]. However, purely vision-based detection often struggles to disambiguate elements that share similar shapes and colors, motivating the integration of additional modalities [23-25].

Recent research has demonstrated that combining image and text features can enhance object detection, particularly in domains where semantic information is embedded in textual labels [26, 27]. For example, Gu et al. [28] proposed a multimodal UI detection framework that integrates OCR text embeddings with visual features, demonstrating improved accuracy over vision-only baselines. More broadly, the integration of visual and textual cues has been explored in multimodal learning, where joint representations capture richer semantics than unimodal approaches [8, 29]. In contrast to OCR-based pipelines, our method leverages GPT-generated descriptions of screenshots, providing higher-level semantic context beyond raw text strings.

Cross-attention has become a central mechanism in vision-language architectures, enabling alignment between image regions and text tokens. Transformer-based models such as ViLBERT [30] and LXMERT [31] introduced co-attention mechanisms to jointly learn from the two modalities, while more recent approaches like MDETR [32] and GLIP [20] incorporate cross-attention into object detection frameworks, effectively grounding textual queries in image regions. These works highlight the effectiveness of attention-based fusion for multimodal reasoning. Building on this insight, our model introduces cross-attention modules into the YOLOv5 neck, allowing GPT-derived text embeddings to dynamically guide feature selection in the detection pipeline. This design provides fine-grained alignment between visual regions and textual descriptions, in contrast to earlier multimodal UI approaches that rely primarily on concatenation or fixed fusion schemes.

Automated software testing has increasingly benefited from advances in computer vision. White et al. [7] integrated YOLO into a GUI testing framework, automatically detecting widgets in desktop applications to improve coverage and robustness. Moran et al. [33] applied vision-based methods to verify mobile app GUIs against design specifications, enabling automated compliance checking. These approaches demonstrate the promise of object detection in testing workflows but remain limited by vision-only perception. Our multimodal method addresses this gap by leveraging textual semantics to improve robustness, reducing false positives in visually ambiguous cases (e.g., distinguishing a "Login" button from a non-interactive label). By combining visual cues with language-based context, our work provides a more reliable perception module for automated testing pipelines.

## 3. Multi-modal detection model

In this section, we first describe the YOLO model, which is the basis of our UI control detection system. We then introduce the cross-attention modules we have developed for merging image and text features. We finally explain the multi-modal UI control detection model resulted from combining the YOLO-based model and cross-attention modules.

### 3.1. User interface detection model

The YOLO detection model [17] is the foundation of our UI control detection system. There are several versions of the YOLO object detection framework, each introducing architectural improvements and performance enhancements over its predecessors. While newer versions such as YOLOv6, YOLOv7, and YOLOv8 offer increased accuracy and speed, they often come with higher computational requirements. In this work, we adopt YOLOv5 as the foundation for our UI control detection system because it strikes an effective balance between detection performance and computational efficiency. Additionally, YOLOv5 remains one of the most widely used and well-supported versions in both research and applied settings, making it a practical choice for integration into our multi-modal model.

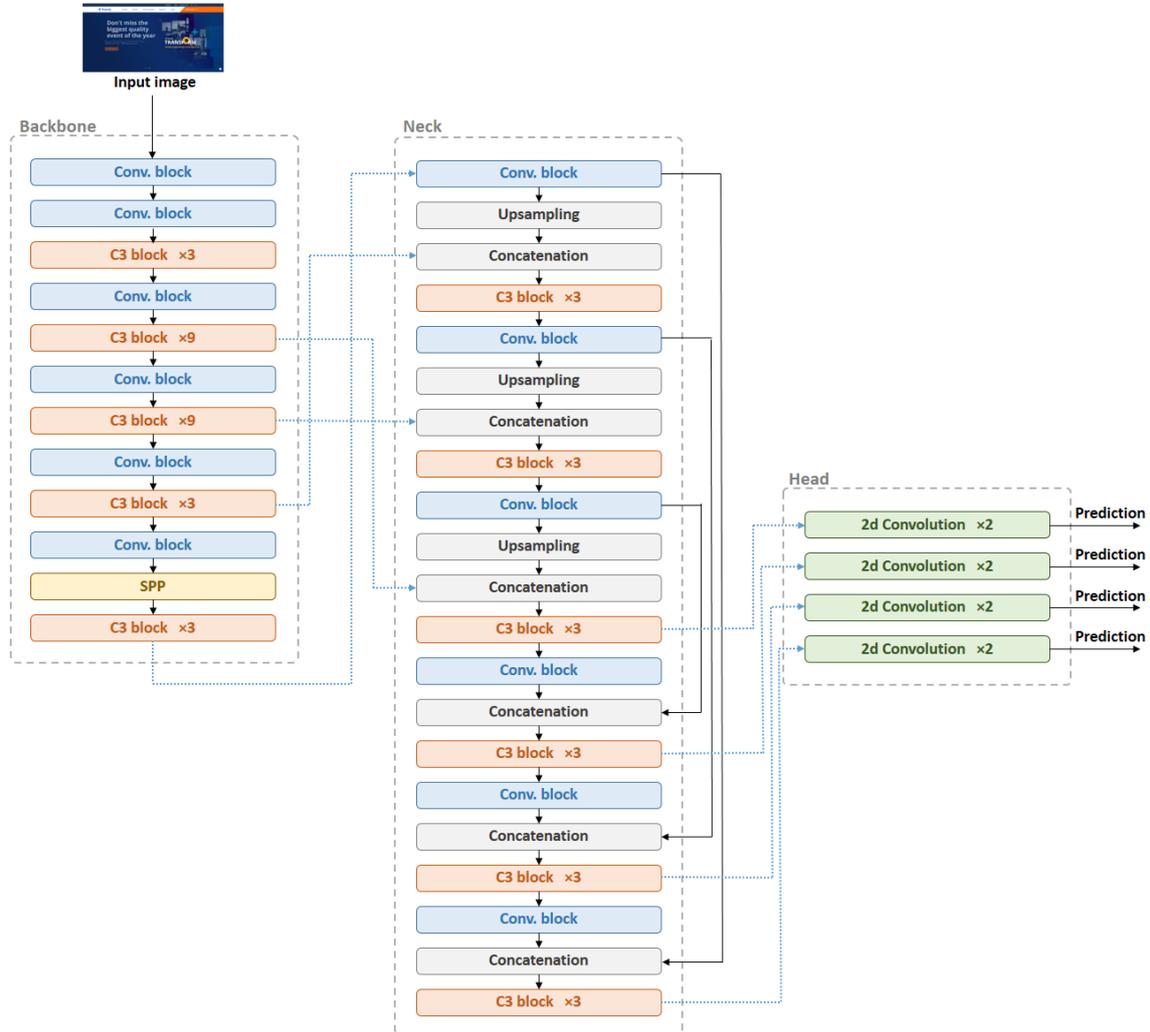

**Figure 1.** The architecture of the YOLOv5 object detection model, which forms the foundation of our UI control detection framework.

The YOLOv5 object detection model is architecturally divided into three main components: the **backbone**, **neck**, and **head**. Figure 1 illustrates the architecture of the YOLO-v5 model. The backbone is responsible for feature extraction and plays a critical role in transforming raw pixel data from an input image into a rich set of hierarchical features. In YOLOv5, the backbone is built on the Cross Stage Partial Network (CSPNet) [34], which improves computational efficiency and feature representation by reducing gradient information duplication during training. The input image is progressively downsampled through convolutional layers and residual connections, capturing both low-level textures and high-level semantic patterns that are essential for recognizing different UI elements.

The neck and head together form the detection path of the YOLOv5 model. The neck serves as an intermediate feature aggregator that enhances multi-scale feature fusion. YOLOv5 uses a Path Aggregation Network (PANet) [35] in the neck to combine features from different levels of the backbone, improving the model's ability to detect UI components of varying sizes and aspect ratios. The head is the final stage that performs object classification and bounding box regression. It generates dense predictions across multiple scales by applying anchor boxes to the aggregated features, outputting coordinates, objectness scores, and class probabilities. By integrating these components into a unified architecture, YOLOv5 enables real-time and accurate object detection from software screenshots, laying the groundwork for the subsequent integration of textual modalities in a multi-modal detection framework.

YOLOv5's backbone and neck are composed of several key architectural blocks that enable efficient and expressive feature extraction. **Convolutional blocks** [36] are the fundamental units, consisting of convolution layers followed by batch normalization and activation functions, used to detect local patterns in the image. **Bottleneck blocks** [37] reduce computational load by compressing and then expanding feature dimensions, preserving important information while improving efficiency. **C3 blocks** (Cross-stage partial bottleneck with 3 convolutions) [38] enhance feature reuse and gradient flow by splitting feature maps, applying bottlenecks to one part, and then merging the outputs. The **Spatial Pyramid Pooling** (SPP) [39] block increases the receptive field and captures multi-scale context by applying pooling operations of different sizes in parallel, which helps in recognizing objects of various scales and spatial arrangements. Together, these blocks form the core of YOLOv5's feature extraction and fusion capabilities. The architectures of these key blocks are illustrated in Figure 2.

### 3.2. Cross-attention modules

To enable joint reasoning over visual and textual modalities, we integrate cross-attention modules into specific locations within the YOLOv5 architecture, particularly after the C3 blocks in the neck. These modules serve as the fusion points where visual features extracted from the image are combined with semantic information derived from textual descriptions. Each cross-attention module receives two inputs: 1) image features from a C3 block, and 2) text features generated by encoding the textual input into a dense embedding. The cross-attention mechanism aligns and weights the image features based on their relevance to the accompanying text [40]. Once attention scores and weights are computed, the resulting attended features must be merged back with the original image features to form a unified multi-modal representation. The way these two feature sets are merged plays a crucial role in the effectiveness and flexibility of the model. In the following paragraphs, we describe three different strategies for merging these features, each implemented as a distinct variant of the cross-attention module.

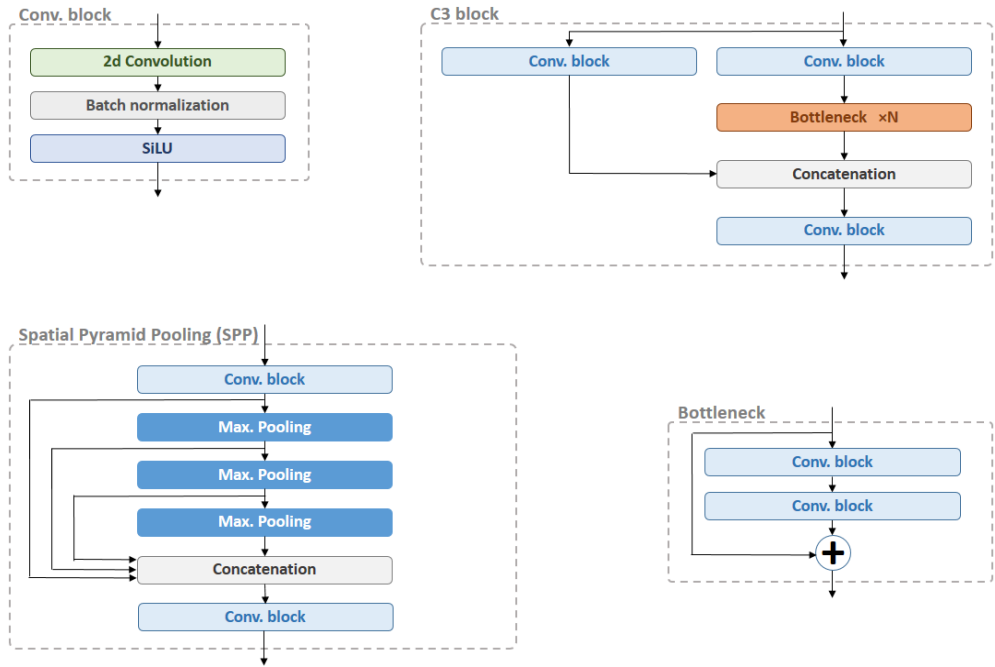

**Figure 2.** Architecture of the main building blocks of the backbone and neck in the YOLOv5 model.

All the cross-attention modules follow a shared computation pattern prior to merging. First, the image features are linearly projected to form the query matrix. Simultaneously, the text features are projected into key and value matrices. Attention scores are then calculated by taking the dot product between the query and key matrices, capturing the similarity between visual and textual tokens. A softmax layer is applied to the attention scores to normalize them into attention weights ranging from 0 to 1. These weights determine how much emphasis each visual feature should place on corresponding textual features. Finally, another matrix multiplication between the attention weights and the value matrix yields the attended features, which encode the textual context aligned to the image regions. The final step is to merge these attended features back with the original image features, which is where different fusion strategies come into play.

The first fusion method is **element-wise addition**, where each element in the attended feature tensor is simply added to the corresponding element in the image feature tensor. This approach is computationally efficient and straightforward to implement, with no additional parameters introduced during merging. It preserves the original spatial structure and maintains alignment between features. However, it may lack flexibility, as it treats both modalities equally and offers no way to control the influence of the text features. This can result in suboptimal fusion when one modality contains more informative signals than the other. Figure 3 illustrates the overall architecture of the cross-attention module based on element-wise addition.

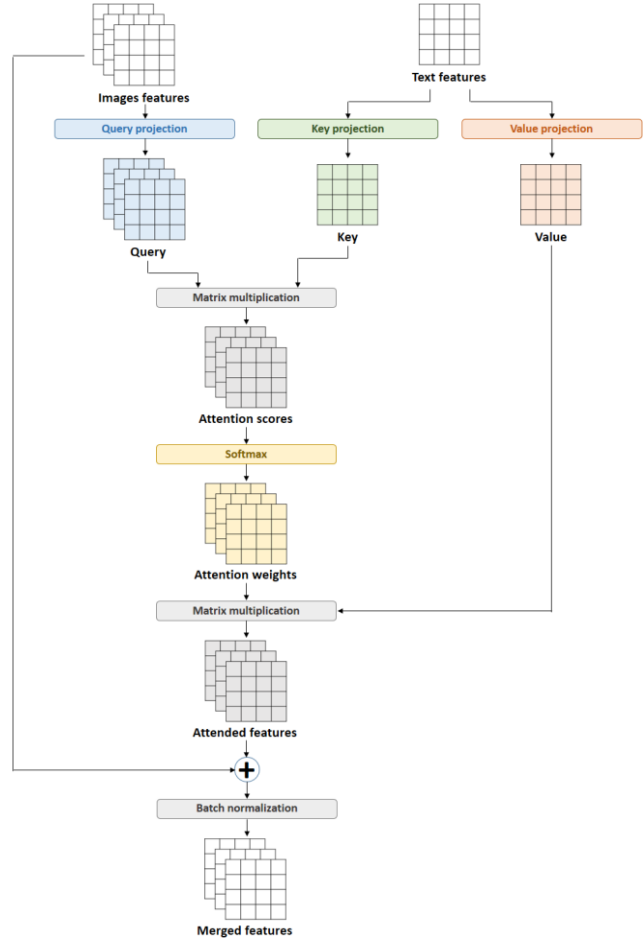

**Figure 3.** The overall architecture and main computations in the cross-attention module based on element-wise addition.

The second fusion method uses a **weighted sum**, where learnable scalar weights are assigned to both the attended and image features before summation. These weights are trained jointly with the rest of the model, allowing it to dynamically adjust the importance of visual and textual inputs during training. This method introduces a small number of additional parameters and offers more control over the fusion process compared to simple addition. The downside may be that it still assumes a uniform weighting across all spatial locations and feature channels, which may limit its expressiveness in capturing fine-grained dependencies between modalities. Figure 4 illustrates the overall architecture of the cross-attention module based on weighted sum.

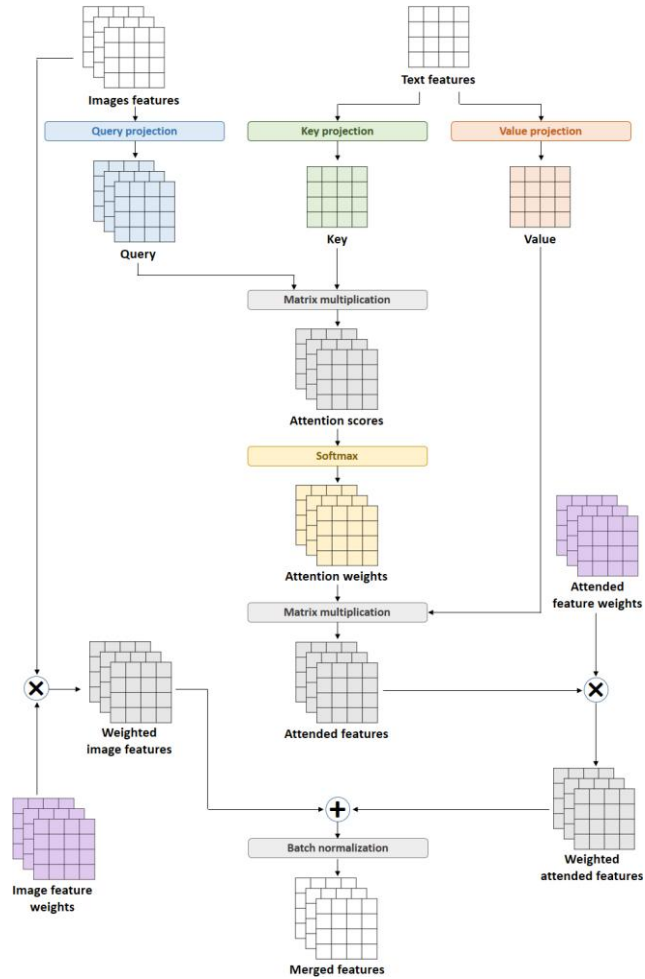

**Figure 4.** The overall architecture and main computations in the cross-attention module based on weighted sum.

The third fusion method is **convolutional fusion**, where the attended and image features are first concatenated along the channel dimension and then passed through a convolutional layer. This approach allows the model to learn complex, non-linear interactions between the two modalities and adaptively integrate them at each spatial location. It is more expressive than the previous methods and leverages the spatial modeling capability of convolutional operations. However, it comes at the cost of increased computational complexity and a higher number of learnable parameters, which may impact inference time and memory usage in resource-constrained environments. Figure 5 illustrates the overall architecture of the cross-attention module based on convolutional fusion.

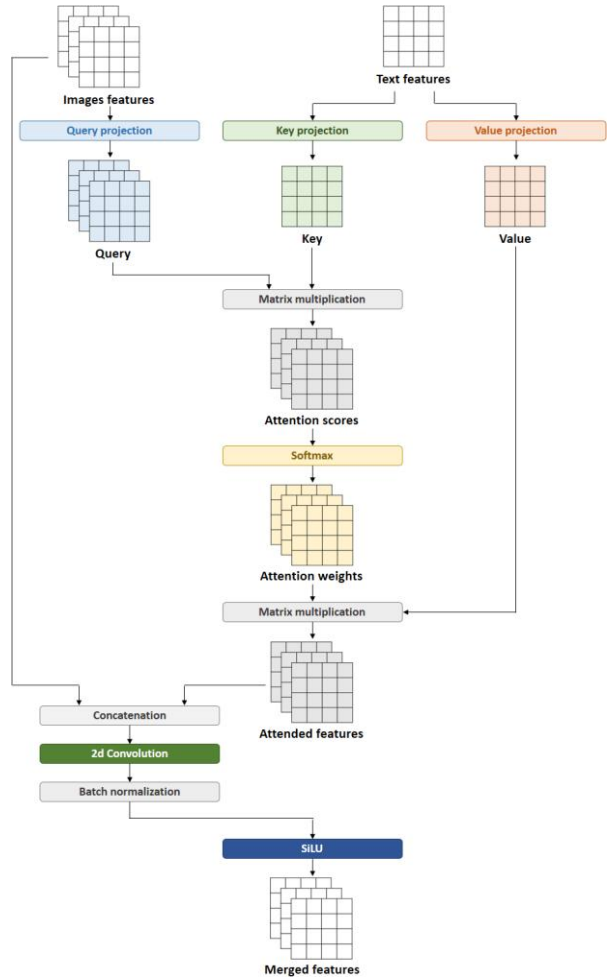

**Figure 5.** The overall architecture and main computations in the cross-attention module based on convolutional fusion.

## 3.3. Multi-modal detection model

Our multi-modal UI control detection model builds upon the YOLOv5 architecture as its foundation, extending it to incorporate textual information through the use of cross-attention modules. These modules are strategically inserted after each C3 block in the neck of the YOLOv5 model, excluding the final C3 block. This design allows the model to progressively integrate semantic cues from the textual description with visual features of varying spatial resolutions and abstraction levels. At each insertion point, a cross-attention module takes the image features produced by the preceding C3 block and fuses them with text features derived from the corresponding image description. This fusion enhances the model's ability to disambiguate and recognize UI elements, especially those that may look visually similar but differ in function or meaning as indicated by accompanying text. Figure 6 illustrates the architecture of the neck in the multimodal detection model after inserting the cross-attention modules. In the multi-modal model, the **backbone**, **head**, and the **connections between the components** remain unchanged from the original YOLOv5 architecture (illustrated in Figure 1), ensuring that the core detection pipeline is preserved while augmenting it with additional semantic information from text.

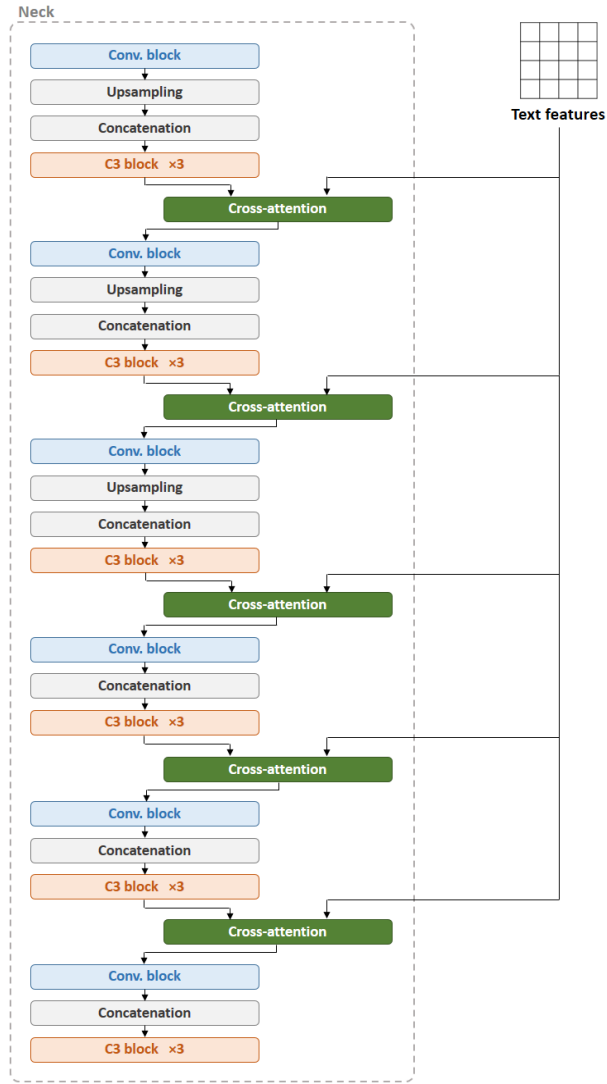

**Figure 6.** The architecture of the neck in the multimodal detection model after inserting the cross-attention modules. In the multi-modal model, the backbone, head, and the connections between the components remain unchanged from the original YOLOv5 architecture.

Placing cross-attention modules after the C3 blocks, as opposed to before, offers a critical advantage: the image features at that stage are already enriched through multi-level residual learning and local-global contextual integration. This allows the attention mechanism to operate on more semantically meaningful features, leading to more effective and relevant fusion with the text embeddings. Furthermore, since the output of the final C3 block is passed directly to the detection heads for final object classification and localization, we omit any cross-attention module after it. Introducing cross-attention at this late stage could disrupt the final prediction logic and add unnecessary computational overhead without providing additional semantic benefit, as the features have already undergone multiple stages of multi-modal enrichment.

To obtain the textual descriptions that guide cross-attention, we use `GPT-4o` [41], a state-of-the-art multi-modal large language model capable of generating high-quality and contextually accurate descriptions of UI images. These descriptions can alternatively be generated manually or semi-automatically in practical settings, depending on application constraints. Once the text is available, it is converted into a dense embedding representation using `text-embedding-3-large`, an advanced embedding model developed by OpenAI. This embedding captures the semantic content of the textual input and serves as the key and value components in the cross-attention computation. While other language models could be used for these steps, we selected these tools for their strong performance and compatibility with our architecture.

We initially trained a standard YOLOv5 model from scratch using a UI control detection dataset, optimizing it solely on visual input. This trained model then served as the starting point for our multi-modal extension. After inserting the cross-attention modules at selected neck locations, we fine-tuned the entire network using pairs of input images and their corresponding text embeddings. This fine-tuning process allowed the model to learn how to jointly reason over both modalities and align them for improved detection accuracy. Further details regarding dataset composition, model training, hyperparameter settings, and evaluation metrics are provided in the experimentation section (Section 4).

Appendix A presents the prompt used for generating textual descriptions from screenshot images by the GPT model, an example of a software screenshot and its textual description generated by the GPT model, and UI controls detected by the multi-modal detection model using the image and textual description as inputs.

## 4. Experimental results

In this section, we first describe the base YOLO model we initially trained on the UI control detection dataset. We then present and discuss experimental results obtained after fine-tuning the multi-modal detection model on both image and text modalities. Furthermore, we discuss the computational complexity of the multi-modal YOLO, and present the ablation study we conducted to investigate the importance of new components in the multi-modal detection model. Source codes, data, and experimentation details are available online at https://github.com/mmoradi-iut/MultiModal-UIDetection.

## 4.1. Baseline model

To establish a baseline for our multi-modal model, we first trained a standard YOLOv5 model on a custom user interface control detection dataset. The dataset contains 16,155 UI screenshots, each annotated with bounding boxes and corresponding class labels. We split the dataset into a training set of 14,155 images and a test set of 2,000 images, with 10% of the training set (1,416 images) used as a validation set during training. The dataset includes 23 distinct UI control classes, representing a wide range of interface components such as buttons, icons, input fields, and charts. All experiments were conducted on a server equipped with a NVIDIA Tesla T4 GPU, 16 processing cores, and 110 GB of RAM.

To evaluate the detection performance of the trained YOLOv5 model, we used four standard object detection metrics: precision, recall, F1-score, and mean Average Precision at IoU threshold 0.5 (mAP@0.5). Precision measures the accuracy of the predicted detections by quantifying how many of them are correct. Recall indicates the ability of the model to detect all relevant objects in the image. The F1-score provides a harmonic mean between precision and recall, offering a balanced measure of the model's performance. Finally, mAP@0.5 evaluates the average precision across all classes at an Intersection-over-Union (IoU) threshold of 0.5, serving as a comprehensive summary of the model's object localization and classification ability.

The YOLOv5 model demonstrated solid overall performance on the test set, as shown in Table 1. The model achieved a precision of 0.765, recall of 0.676, F1-score of 0.701, and mAP@0.5 of 0.649 across all 23 classes. The strongest results were observed for frequently occurring and visually distinctive controls such as *Image* (F1: 0.882, mAP: 0.820), *Button* (F1: 0.874, mAP: 0.861), and *Radio_Selected* (F1: 0.875, mAP: 0.806). Similarly, *Text*, *Table*, and *Checkbox_Unchecked* also achieved high scores across all metrics. On the other hand, the model struggled with more abstract or spatially ambiguous elements like *Horizontal_Axis* (F1: 0.0736, mAP: 0.0128), *Vertical_Axis* (F1: 0.204, mAP: 0.125), and *Label_of_the_Textarea* (F1: 0.622, mAP: 0.476). These results suggest that while the model effectively detects most common UI components, it faces challenges with less prominent or semantically complex elements, motivating the need for incorporating complementary textual information in the multi-modal approach.

## 4.2. Multi-modal model

In this subsection, we present and discuss the performance scores obtained by the multi-modal UI detection model, separately for each cross-attention merging method described in Section 3.2.

**Table 1.** Performance scores obtained by the baseline YOLO model on the user interface control detection dataset.

| Class | Precision | Recall | F1-score | mAP@.5 |
|---|---|---|---|---|
| all | 0.765 | 0.676 | 0.701 | 0.649 |
| Icon | 0.924 | 0.773 | 0.842 | 0.766 |
| Dropdown | 0.842 | 0.808 | 0.825 | 0.788 |
| Button | 0.848 | 0.902 | 0.874 | 0.861 |
| Menu | 0.750 | 0.530 | 0.621 | 0.507 |
| Input | 0.859 | 0.762 | 0.808 | 0.757 |
| List | 0.613 | 0.782 | 0.687 | 0.658 |
| TabBar | 0.857 | 0.588 | 0.697 | 0.601 |
| Table | 0.788 | 0.839 | 0.813 | 0.796 |
| Radio_Selected | 0.939 | 0.820 | 0.875 | 0.806 |
| Radio_Unselected | 0.793 | 0.907 | 0.846 | 0.813 |
| Checkbox_Unchecked | 0.913 | 0.825 | 0.866 | 0.836 |
| Checkbox_Checked | 0.849 | 0.581 | 0.690 | 0.566 |
| Tree | 0.729 | 0.788 | 0.758 | 0.738 |
| Image | 0.928 | 0.839 | 0.882 | 0.820 |
| Text | 0.928 | 0.797 | 0.857 | 0.802 |
| Label_of_the_Textarea | 0.896 | 0.476 | 0.622 | 0.476 |
| Description_List | 0.700 | 0.801 | 0.747 | 0.703 |
| Legend | 0.800 | 0.647 | 0.715 | 0.647 |
| Horizontal_Axis | 0.117 | 0.053 | 0.073 | 0.012 |
| Chart | 0.603 | 0.755 | 0.671 | 0.715 |
| Graph | 0.607 | 0.575 | 0.590 | 0.572 |
| Vertical_Axis | 0.760 | 0.118 | 0.204 | 0.125 |
| Date_area | 0.557 | 0.588 | 0.572 | 0.567 |

### 4.2.1. Element-wise addition

The evaluation results of the multi-modal model incorporating element-wise addition for merging image and attended text features show a consistent improvement over the baseline YOLOv5 model. As seen in Table 2, increasing the number of cross-attention modules from three to five leads to steady gains in all core metrics. The model with five cross-attention blocks achieved an overall precision of 0.791, recall of 0.684, F1-score of 0.719, and mAP@0.5 of 0.681, outperforming the baseline model. Even with just three cross-attention blocks, the model already matched or slightly exceeded the baseline in several metrics. This trend clearly demonstrates the benefit of integrating textual semantic information, even using the simplest form of merging via element-wise addition.

When comparing per-class performance, we observe that categories previously underperforming in the baseline model benefited from the multi-modal architecture. For instance, *Horizontal_Axis*, which had very low detection performance in the baseline (F1: 0.0736,

mAP@0.0128), showed significant improvement with five cross-attention modules (F1: 0.160, mAP@0.265). Other categories like *Label_of_the_Textarea*, *Checkbox_Checked*, and *Legend* also showed notable F1-score and mAP gains. While high-performing classes such as *Button*, *Image*, and *Text* remained relatively stable, the introduction of text-aware fusion helped elevate the performance of more ambiguous or sparsely represented classes. Overall, the results validate that even simple fusion techniques in a multi-modal setup can significantly improve detection performance, especially for semantically complex UI elements.

**Table 2.** Performance scores obtained by the multi-modal UI detection model incorporating element-wise addition for merging image and attended text features.

| Class | Precision | | | Recall | | | F1-score | | | mAP@.5 | | |
|---|---|---|---|---|---|---|---|---|---|---|---|---|
| | Cross-attention blocks | | | Cross-attention blocks | | | Cross-attention blocks | | | Cross-attention blocks | | |
| | 3 | 4 | 5 | 3 | 4 | 5 | 3 | 4 | 5 | 3 | 4 | 5 |
| all | 0.761 | 0.777 | 0.791 | 0.669 | 0.675 | 0.684 | 0.693 | 0.707 | 0.719 | 0.651 | 0.665 | 0.681 |
| Icon | 0.920 | 0.926 | 0.924 | 0.767 | 0.767 | 0.784 | 0.836 | 0.839 | 0.848 | 0.768 | 0.783 | 0.792 |
| Dropdown | 0.852 | 0.866 | 0.823 | 0.751 | 0.767 | 0.741 | 0.798 | 0.813 | 0.779 | 0.737 | 0.748 | 0.767 |
| Button | 0.845 | 0.853 | 0.852 | 0.900 | 0.895 | 0.902 | 0.871 | 0.873 | 0.876 | 0.864 | 0.878 | 0.882 |
| Menu | 0.794 | 0.807 | 0.819 | 0.522 | 0.542 | 0.571 | 0.629 | 0.648 | 0.672 | 0.532 | 0.545 | 0.562 |
| Input | 0.845 | 0.856 | 0.864 | 0.759 | 0.765 | 0.775 | 0.799 | 0.807 | 0.817 | 0.752 | 0.762 | 0.748 |
| List | 0.621 | 0.631 | 0.639 | 0.734 | 0.754 | 0.776 | 0.672 | 0.687 | 0.700 | 0.628 | 0.647 | 0.647 |
| TabBar | 0.877 | 0.888 | 0.877 | 0.589 | 0.617 | 0.632 | 0.704 | 0.728 | 0.734 | 0.579 | 0.582 | 0.616 |
| Table | 0.763 | 0.803 | 0.791 | 0.837 | 0.825 | 0.841 | 0.798 | 0.813 | 0.815 | 0.775 | 0.786 | 0.764 |
| Radio_Selected | 0.908 | 0.918 | 0.929 | 0.833 | 0.812 | 0.836 | 0.868 | 0.861 | 0.880 | 0.830 | 0.842 | 0.857 |
| Radio_Unselected | 0.784 | 0.768 | 0.818 | 0.894 | 0.896 | 0.882 | 0.835 | 0.827 | 0.848 | 0.827 | 0.819 | 0.836 |
| Checkbox_Unchecked | 0.901 | 0.907 | 0.914 | 0.794 | 0.794 | 0.814 | 0.844 | 0.846 | 0.861 | 0.834 | 0.834 | 0.846 |
| Checkbox_Checked | 0.815 | 0.824 | 0.831 | 0.649 | 0.685 | 0.694 | 0.722 | 0.748 | 0.756 | 0.603 | 0.625 | 0.726 |
| Tree | 0.733 | 0.792 | 0.709 | 0.794 | 0.801 | 0.783 | 0.762 | 0.796 | 0.744 | 0.748 | 0.754 | 0.711 |
| Image | 0.942 | 0.947 | 0.936 | 0.862 | 0.856 | 0.765 | 0.900 | 0.899 | 0.841 | 0.846 | 0.859 | 0.859 |
| Text | 0.925 | 0.932 | 0.935 | 0.795 | 0.798 | 0.796 | 0.855 | 0.859 | 0.859 | 0.807 | 0.814 | 0.807 |
| Label_of_the_Textarea | 0.876 | 0.885 | 0.897 | 0.421 | 0.453 | 0.476 | 0.568 | 0.599 | 0.621 | 0.412 | 0.447 | 0.481 |
| Description_List | 0.695 | 0.725 | 0.742 | 0.775 | 0.767 | 0.797 | 0.732 | 0.745 | 0.768 | 0.691 | 0.699 | 0.704 |
| Legend | 0.800 | 0.804 | 0.811 | 0.647 | 0.672 | 0.686 | 0.715 | 0.732 | 0.743 | 0.585 | 0.627 | 0.679 |
| Horizontal_Axis | 0.098 | 0.042 | 0.225 | 0.0362 | 0.053 | 0.125 | 0.053 | 0.046 | 0.160 | 0.188 | 0.220 | 0.265 |
| Chart | 0.678 | 0.735 | 0.636 | 0.745 | 0.702 | 0.714 | 0.709 | 0.718 | 0.672 | 0.700 | 0.698 | 0.714 |
| Graph | 0.590 | 0.631 | 0.656 | 0.600 | 0.600 | 0.625 | 0.594 | 0.615 | 0.640 | 0.565 | 0.580 | 0.597 |
| Vertical_Axis | 0.818 | 0.844 | 0.803 | 0.099 | 0.130 | 0.142 | 0.177 | 0.225 | 0.241 | 0.147 | 0.179 | 0.213 |
| Date_area | 0.436 | 0.500 | 0.780 | 0.588 | 0.588 | 0.588 | 0.500 | 0.540 | 0.670 | 0.571 | 0.588 | 0.592 |

### 4.2.2. Weighted sum

The multi-modal model that uses weighted sum for merging image and attended text features outperforms both the baseline YOLOv5 and the element-wise addition variant in nearly all key metrics. As shown in Table 3, the best-performing version (with five cross-attention modules) achieved an overall precision of 0.805, recall of 0.701, F1-score of 0.738, and mAP@0.5 of 0.703.

These results represent a significant improvement over the baseline (F1: 0.701, mAP@0.5: 0.649) and a noticeable gain over the element-wise addition model (F1: 0.719, mAP@0.5: 0.681). The improvement is consistent as the number of cross-attention blocks increases, highlighting the effectiveness of incorporating learned weights during feature fusion. The weighted sum allows the model to adaptively balance visual and textual contributions, leading to better alignment and richer multi-modal representations.

**Table 3.** Performance scores obtained by the multi-modal UI detection model incorporating weighted sum for merging image and attended text features.

| Class | Precision | | | Recall | | | F1-score | | | mAP@.5 | | |
|---|---|---|---|---|---|---|---|---|---|---|---|---|
| | Cross-attention blocks | | | Cross-attention blocks | | | Cross-attention blocks | | | Cross-attention blocks | | |
| | 3 | 4 | 5 | 3 | 4 | 5 | 3 | 4 | 5 | 3 | 4 | 5 |
| all | 0.771 | 0.781 | 0.805 | 0.671 | 0.687 | 0.701 | 0.702 | 0.718 | 0.738 | 0.654 | 0.680 | 0.703 |
| Icon | 0.924 | 0.929 | 0.933 | 0.768 | 0.774 | 0.784 | 0.838 | 0.844 | 0.852 | 0.773 | 0.782 | 0.809 |
| Dropdown | 0.859 | 0.866 | 0.852 | 0.687 | 0.751 | 0.765 | 0.763 | 0.804 | 0.806 | 0.719 | 0.743 | 0.757 |
| Button | 0.847 | 0.854 | 0.860 | 0.897 | 0.909 | 0.902 | 0.871 | 0.880 | 0.880 | 0.863 | 0.899 | 0.874 |
| Menu | 0.805 | 0.815 | 0.827 | 0.520 | 0.538 | 0.559 | 0.631 | 0.648 | 0.667 | 0.529 | 0.540 | 0.572 |
| Input | 0.852 | 0.864 | 0.855 | 0.763 | 0.762 | 0.772 | 0.805 | 0.809 | 0.811 | 0.757 | 0.756 | 0.764 |
| List | 0.645 | 0.676 | 0.697 | 0.737 | 0.745 | 0.766 | 0.687 | 0.708 | 0.729 | 0.651 | 0.667 | 0.694 |
| TabBar | 0.896 | 0.907 | 0.915 | 0.581 | 0.594 | 0.620 | 0.704 | 0.717 | 0.739 | 0.578 | 0.613 | 0.641 |
| Table | 0.812 | 0.821 | 0.834 | 0.831 | 0.836 | 0.839 | 0.821 | 0.828 | 0.836 | 0.768 | 0.796 | 0.807 |
| Radio_Selected | 0.913 | 0.913 | 0.926 | 0.837 | 0.846 | 0.845 | 0.873 | 0.878 | 0.883 | 0.836 | 0.833 | 0.833 |
| Radio_Unselected | 0.750 | 0.764 | 0.784 | 0.896 | 0.899 | 0.908 | 0.816 | 0.826 | 0.841 | 0.815 | 0.823 | 0.841 |
| Checkbox_Unchecked | 0.910 | 0.916 | 0.907 | 0.778 | 0.799 | 0.808 | 0.838 | 0.853 | 0.854 | 0.834 | 0.850 | 0.840 |
| Checkbox_Checked | 0.836 | 0.829 | 0.849 | 0.632 | 0.653 | 0.665 | 0.719 | 0.730 | 0.745 | 0.611 | 0.634 | 0.676 |
| Tree | 0.753 | 0.759 | 0.780 | 0.788 | 0.794 | 0.808 | 0.770 | 0.776 | 0.793 | 0.741 | 0.760 | 0.760 |
| Image | 0.944 | 0.943 | 0.955 | 0.862 | 0.868 | 0.874 | 0.901 | 0.903 | 0.912 | 0.852 | 0.852 | 0.869 |
| Text | 0.930 | 0.925 | 0.937 | 0.797 | 0.804 | 0.810 | 0.858 | 0.860 | 0.868 | 0.812 | 0.845 | 0.845 |
| Label_of_the_Textarea | 0.882 | 0.873 | 0.888 | 0.397 | 0.433 | 0.433 | 0.547 | 0.578 | 0.582 | 0.454 | 0.491 | 0.517 |
| Description_List | 0.738 | 0.725 | 0.740 | 0.783 | 0.779 | 0.779 | 0.759 | 0.751 | 0.758 | 0.706 | 0.738 | 0.747 |
| Legend | 0.800 | 0.800 | 0.833 | 0.647 | 0.647 | 0.647 | 0.715 | 0.715 | 0.728 | 0.608 | 0.635 | 0.665 |
| Horizontal_Axis | 0.098 | 0.218 | 0.438 | 0.213 | 0.291 | 0.358 | 0.134 | 0.249 | 0.393 | 0.190 | 0.223 | 0.381 |
| Chart | 0.689 | 0.698 | 0.712 | 0.702 | 0.752 | 0.752 | 0.695 | 0.723 | 0.731 | 0.677 | 0.741 | 0.734 |
| Graph | 0.612 | 0.584 | 0.675 | 0.592 | 0.600 | 0.621 | 0.601 | 0.591 | 0.646 | 0.563 | 0.592 | 0.607 |
| Vertical_Axis | 0.815 | 0.827 | 0.834 | 0.198 | 0.218 | 0.265 | 0.318 | 0.345 | 0.402 | 0.236 | 0.295 | 0.371 |
| Date_area | 0.431 | 0.458 | 0.497 | 0.529 | 0.529 | 0.543 | 0.474 | 0.490 | 0.518 | 0.490 | 0.541 | 0.575 |

Comparing per-class metrics further illustrates the advantage of the weighted sum approach. Notably, previously challenging classes like *Horizontal_Axis* saw substantial gains, reaching an F1-score of 0.393 and mAP@0.5 of 0.381 with five cross-attention modules, far exceeding both the baseline and element-wise models. Similarly, improvements were observed in classes like *Checkbox_Checked*, *Vertical_Axis*, and *Graph*, where the added flexibility in the fusion process helped recover subtle semantic cues. For common and well-defined classes such as *Image*, *Button*, and *Text*, the model maintained or slightly improved its already high performance. Overall, these results demonstrate that the weighted sum fusion method not only enhances general detection quality but also better addresses the weaknesses seen in the baseline and simpler fusion models, making it a more robust multi-modal solution.

### 4.2.3. Convolutional fusion

The multi-modal model that uses convolutional fusion to merge image and attended text features achieves the best overall performance among all models evaluated in this study. As shown in Table 4, the model with five cross-attention modules reaches an overall precision of 0.820, recall of 0.725, F1-score of 0.761, and mAP@0.5 of 0.732. These metrics represent the highest values across all models, surpassing the baseline YOLOv5 (F1: 0.701, mAP@0.5: 0.649), the element-wise addition variant (F1: 0.719, mAP@0.5: 0.681), and the weighted sum variant (F1: 0.738, mAP@0.5: 0.703). The use of convolutional layers for fusion allows the model to learn complex, non-linear interactions between the visual and textual modalities at each spatial location, which contributes to these superior results.

The advantages of convolutional fusion are especially visible in more semantically complex or visually ambiguous classes. For instance, *Horizontal_Axis* (a particularly challenging class) achieves an F1-score of 0.468 and mAP@0.5 of 0.509, which are significantly higher than what any other model configuration achieved. Similar improvements are observed for *Label_of_the_Textarea*, *Chart*, *Graph*, and *Vertical_Axis*, all of which benefit from the spatial and semantic refinement enabled by the convolutional merging. While this method introduces more computational overhead due to the added convolutional layers, the trade-off is justified by the consistently better performance across the board. The results highlight convolutional fusion as the most effective approach for multi-modal UI control detection, particularly in scenarios where fine-grained visual-textual alignment is critical.

### 4.3. Ablation study

To complement the comprehensive evaluation of our multi-modal UI control detection model, this subsection presents the results of an ablation study designed to isolate and examine the impact of key architectural and input design choices. As detailed in previous sections, our paper already includes an extensive set of ablation experiments that investigate the performance effects of three fusion strategies, i.e. element-wise addition, weighted sum, and convolutional fusion, alongside varying the number of cross-attention modules (three, four, and five). These experiments established a clear understanding of how different configurations of our model architecture contribute to detection performance.

**Table 4.** Performance scores obtained by the multi-modal UI detection model incorporating convolutional fusion for merging image and attended text features.

| Class | Precision | | | Recall | | | F1-score | | | mAP@.5 | | |
|---|---|---|---|---|---|---|---|---|---|---|---|---|
| | Cross-attention blocks | | | Cross-attention blocks | | | Cross-attention blocks | | | Cross-attention blocks | | |
| | 3 | 4 | 5 | 3 | 4 | 5 | 3 | 4 | 5 | 3 | 4 | 5 |
| all | 0.790 | 0.801 | 0.820 | 0.682 | 0.707 | 0.725 | 0.715 | 0.740 | 0.761 | 0.680 | 0.711 | 0.732 |
| Icon | 0.915 | 0.928 | 0.928 | 0.781 | 0.797 | 0.788 | 0.842 | 0.857 | 0.852 | 0.784 | 0.790 | 0.784 |
| Dropdown | 0.856 | 0.856 | 0.862 | 0.791 | 0.800 | 0.796 | 0.822 | 0.827 | 0.827 | 0.762 | 0.791 | 0.791 |
| Button | 0.850 | 0.864 | 0.864 | 0.908 | 0.908 | 0.910 | 0.878 | 0.885 | 0.886 | 0.871 | 0.872 | 0.872 |
| Menu | 0.793 | 0.793 | 0.825 | 0.535 | 0.568 | 0.584 | 0.638 | 0.661 | 0.683 | 0.548 | 0.581 | 0.590 |
| Input | 0.834 | 0.847 | 0.847 | 0.767 | 0.792 | 0.773 | 0.799 | 0.818 | 0.808 | 0.757 | 0.762 | 0.762 |
| List | 0.674 | 0.698 | 0.731 | 0.749 | 0.773 | 0.754 | 0.709 | 0.733 | 0.742 | 0.671 | 0.696 | 0.744 |
| TabBar | 0.858 | 0.862 | 0.862 | 0.595 | 0.638 | 0.661 | 0.702 | 0.733 | 0.748 | 0.580 | 0.632 | 0.632 |
| Table | 0.779 | 0.796 | 0.823 | 0.841 | 0.857 | 0.869 | 0.808 | 0.825 | 0.845 | 0.786 | 0.792 | 0.817 |
| Radio_Selected | 0.890 | 0.890 | 0.890 | 0.817 | 0.817 | 0.817 | 0.851 | 0.851 | 0.851 | 0.804 | 0.804 | 0.827 |
| Radio_Unselected | 0.828 | 0.828 | 0.835 | 0.907 | 0.907 | 0.907 | 0.865 | 0.865 | 0.869 | 0.889 | 0.889 | 0.889 |
| Checkbox_Unchecked | 0.898 | 0.901 | 0.901 | 0.811 | 0.832 | 0.832 | 0.852 | 0.865 | 0.865 | 0.809 | 0.812 | 0.831 |
| Checkbox_Checked | 0.837 | 0.848 | 0.866 | 0.682 | 0.707 | 0.738 | 0.751 | 0.771 | 0.796 | 0.667 | 0.719 | 0.719 |
| Tree | 0.761 | 0.779 | 0.795 | 0.805 | 0.805 | 0.805 | 0.782 | 0.791 | 0.799 | 0.759 | 0.782 | 0.818 |
| Image | 0.939 | 0.939 | 0.947 | 0.878 | 0.880 | 0.889 | 0.907 | 0.908 | 0.917 | 0.861 | 0.873 | 0.873 |
| Text | 0.920 | 0.932 | 0.932 | 0.812 | 0.815 | 0.827 | 0.862 | 0.869 | 0.876 | 0.815 | 0.839 | 0.839 |
| Label_of_the_Textarea | 0.886 | 0.892 | 0.892 | 0.443 | 0.490 | 0.535 | 0.590 | 0.632 | 0.668 | 0.505 | 0.556 | 0.556 |
| Description_List | 0.692 | 0.709 | 0.753 | 0.795 | 0.795 | 0.825 | 0.739 | 0.749 | 0.787 | 0.701 | 0.701 | 0.724 |
| Legend | 0.800 | 0.812 | 0.838 | 0.552 | 0.582 | 0.582 | 0.653 | 0.678 | 0.686 | 0.610 | 0.651 | 0.687 |
| Horizontal_Axis | 0.410 | 0.459 | 0.497 | 0.174 | 0.289 | 0.443 | 0.244 | 0.354 | 0.468 | 0.381 | 0.418 | 0.509 |
| Chart | 0.737 | 0.745 | 0.789 | 0.739 | 0.788 | 0.788 | 0.737 | 0.765 | 0.788 | 0.709 | 0.760 | 0.775 |
| Graph | 0.687 | 0.698 | 0.757 | 0.575 | 0.592 | 0.625 | 0.626 | 0.640 | 0.684 | 0.562 | 0.619 | 0.674 |
| Vertical_Axis | 0.854 | 0.854 | 0.878 | 0.161 | 0.258 | 0.319 | 0.270 | 0.396 | 0.467 | 0.249 | 0.381 | 0.456 |
| Date_area | 0.476 | 0.497 | 0.556 | 0.588 | 0.588 | 0.626 | 0.526 | 0.538 | 0.588 | 0.578 | 0.638 | 0.673 |

To further strengthen this analysis, we conducted two additional ablation techniques aimed at assessing the role of textual input quality. The first is **mismatched text ablation**, where we replace accurate textual descriptions with unrelated or incorrect ones to evaluate how performance deteriorates under misleading semantic context. The second is **partial text ablation**, where we remove mentions of specific UI controls from otherwise correct descriptions to analyze the impact of incomplete information. These tests allow us to benchmark the influence of textual alignment on detection accuracy, demonstrating that accurate and contextually aligned descriptions significantly enhance performance, while mismatched or incomplete text degrades it, quantifying the model's dependence on high-quality language input.

### 4.3.1. Mismatched text ablation

For the mismatched text ablation, we evaluated how the multi-modal model responds to deliberately incorrect textual descriptions. In this setup, we took the textual description associated with each image and modified it by randomly replacing the class label of every UI control with a different, incorrect label. We also altered location references within the descriptions to ensure that both the semantic and spatial cues were misleading. These mismatched descriptions were generated using GPT-4o, allowing us to automate the ablation while maintaining syntactically plausible but semantically incorrect input. We then evaluated the best-performing multi-modal model, i.e. the one using convolutional fusion with five cross-attention modules, on this corrupted dataset. The results are presented in Table 5, which also includes the percentage of change in performance compared to the baseline YOLOv5 model.

**Table 5.** Performance scores obtained by the best multi-modal model on mismatched text ablations. The percentage of change in scores compared to the baseline model is presented in parenthesis for each score.

| Class | Precision | Recall | F1-score | mAP@.5 |
|---|---|---|---|---|
| all | 0.661 (−13.5%) | 0.581 (−14.0%) | 0.605 (−13.7%) | 0.554 (−14.5%) |
| Icon | 0.771 (−16.5%) | 0.643 (−16.8%) | 0.701 (−16.7%) | 0.657 (−14.2%) |
| Dropdown | 0.738 (−12.3%) | 0.685 (−15.2%) | 0.710 (−13.8%) | 0.635 (−19.4%) |
| Button | 0.725 (−14.5%) | 0.764 (−15.2%) | 0.743 (−14.8%) | 0.711 (−17.4%) |
| Menu | 0.684 (−8.8%) | 0.499 (−5.8%) | 0.577 (−7.0%) | 0.451 (−11.0%) |
| Input | 0.706 (−17.8%) | 0.660 (−13.3%) | 0.682 (−15.5%) | 0.608 (−19.6%) |
| List | 0.530 (−13.5%) | 0.611 (−21.8%) | 0.567 (−17.3%) | 0.553 (−15.9%) |
| TabBar | 0.659 (−23.1%) | 0.537 (−8.6%) | 0.591 (−15.0%) | 0.490 (−18.4%) |
| Table | 0.672 (−14.7%) | 0.708 (−15.6%) | 0.689 (−15.1%) | 0.638 (−19.8%) |
| Radio_Selected | 0.675 (−28.1%) | 0.675 (−17.6%) | 0.675 (−22.8%) | 0.702 (−12.9%) |
| Radio_Unselected | 0.715 (−9.8%) | 0.758 (−16.4%) | 0.735 (−13.0%) | 0.754 (−7.2%) |
| Checkbox_Unchecked | 0.701 (−23.2%) | 0.707 (−14.3%) | 0.703 (−18.7%) | 0.695 (−16.8%) |
| Checkbox_Checked | 0.744 (−12.3%) | 0.530 (−8.7%) | 0.619 (−10.2%) | 0.529 (−6.5%) |
| Tree | 0.669 (−8.2%) | 0.712 (−9.6%) | 0.689 (−8.9%) | 0.633 (−14.2%) |
| Image | 0.815 (−12.1%) | 0.714 (−14.8%) | 0.761 (−13.7%) | 0.698 (−14.8%) |
| Text | 0.802 (−13.5%) | 0.706 (−11.4%) | 0.750 (−12.3%) | 0.739 (−7.8%) |
| Label_of_the_Textarea | 0.750 (−16.2%) | 0.394 (−17.2%) | 0.516 (−16.9%) | 0.395 (−17.0%) |
| Description_List | 0.603 (−13.8%) | 0.745 (−6.9%) | 0.666 (−10.7%) | 0.565 (−19.6%) |
| Legend | 0.672 (−16.0%) | 0.491 (−24.1%) | 0.567 (−20.6%) | 0.477 (−26.2%) |
| Horizontal_Axis | 0.215 (+83.7%) | 0.145 (+170.0%) | 0.173 (+135.3%) | 0.174 (+1259.3%) |
| Chart | 0.633 (+4.9%) | 0.623 (−17.4%) | 0.627 (−6.4%) | 0.610 (−14.6%) |
| Graph | 0.590 (−2.8%) | 0.458 (−20.3%) | 0.515 (−12.5%) | 0.415 (−27.4%) |
| Vertical_Axis | 0.745 (−1.9%) | 0.127 (+7.6%) | 0.217 (+6.3%) | 0.209 (+67.2%) |
| Date_area | 0.394 (−29.2%) | 0.485 (−17.5%) | 0.434 (−23.9%) | 0.416 (−26.6%) |

The results show a clear degradation in performance across all evaluation metrics when comparing the model on mismatched text to the same model using correct text descriptions. Compared to the baseline model, the mismatched-text version also shows performance drops on nearly all class labels, with a few exceptions. Interestingly, scores on classes such as *Horizontal_Axis*, *Chart*, and *Vertical_Axis* are still better than those of the baseline, which come from the significant boost achieved by the multi-modal fine-tuning. Overall, compared to the baseline model, the average precision dropped by 13.5%, recall by 14.0%, F1-score by 13.7%, and mAP@0.5 by 14.5%, clearly demonstrating the disruptive effect of inaccurate textual context.

This degradation highlights two key vulnerabilities of the multi-modal model. First, the drop in precision and recall suggests that incorrect text can lead the model to misclassify controls or completely miss them, disrupting both class prediction and object presence detection. Second, the decrease in mAP@0.5 underscores how inaccurate descriptions can affect the model's spatial reasoning, resulting in poor localization and bounding box quality. These findings confirm that while the model successfully learns to align image and text features during fine-tuning, it also becomes highly dependent on accurate textual context. As such, the multi-modal model is not yet robust to noisy or misleading language input, pointing to a need for future work on improving resilience, possibly through techniques like noise-aware training or confidence-based text filtering.

### 4.3.2. Partial text ablation

For the partial text ablation, we aimed to assess the importance of individual class mentions within the textual descriptions. Specifically, for each class label, we removed all references to that particular UI control from the textual descriptions while keeping the rest of the text intact. This process was repeated separately for every class. The resulting partial descriptions were then used to test the best multi-modal model, i.e. the one employing convolutional fusion with five cross-attention layers, to isolate the effect of missing textual information related to specific classes. The results for each class label are reported in Table 6, along with the percentage of change in the scores relative to the baseline YOLOv5 model.

The results show consistent performance degradation across all metrics and class labels compared to the original multi-modal model with complete textual descriptions. In nearly every class, the absence of specific class references led to a decline in all four evaluation measures. Compared to the baseline model, most classes still experienced a performance drop, with only a few still showing better scores than the baseline. These exceptions, such as *Horizontal_Axis* and *Vertical_Axis*, are likely due to the model initially having a large boost on these rare classes during fine-tuning with text, meaning their removal had a slightly negative effect, but still performing better than the baseline. The overall pattern indicates that recall suffered more than precision, suggesting that when class mentions are missing from the description, the model is more likely to miss relevant UI controls altogether, though it can still often correctly classify the ones it detects. Additionally, the mAP@0.5 drop across most classes highlights a loss in localization accuracy, reflecting how textual cues also support the model's ability to generate precise bounding boxes.

These findings reinforce the conclusion that the multi-modal model successfully learns to align and leverage image-text representations during training, but also reveal a sensitivity to missing textual information. The reliance on class-specific text, while beneficial under normal conditions, becomes a liability when such references are absent. This suggests that although multi-modal training can significantly boost performance, the model should be further reinforced, perhaps via regularization, robustness training, or uncertainty modeling, to maintain competitive accuracy even in the absence of complete textual clues. Such improvements would ensure that the model does not underperform relative to the baseline in real-world settings where textual input may be noisy or incomplete.

**Table 6.** Performance scores obtained by the best multi-modal model on partial text ablations. The percentage of change in scores compared to the baseline model is presented in parenthesis for each score.

| Class | Precision | Recall | F1-score | mAP@.5 |
| --- | --- | --- | --- | --- |
| Icon | 0.855 (−7.4%) | 0.668 (−13.5%) | 0.750 (−10.9%) | 0.728 (−4.9%) |
| Dropdown | 0.778 (−7.6%) | 0.725 (−10.2%) | 0.750 (−8.9%) | 0.735 (−6.7%) |
| Button | 0.792 (−6.6%) | 0.814 (−9.7%) | 0.802 (−8.1%) | 0.810 (−5.9%) |
| Menu | 0.711 (−5.2%) | 0.480 (−9.4%) | 0.573 (−7.7%) | 0.472 (−6.9%) |
| Input | 0.816 (−5.0%) | 0.703 (−7.7%) | 0.755 (−6.4%) | 0.725 (−4.2%) |
| List | 0.587 (−4.2%) | 0.699 (−10.6%) | 0.638 (−7.1%) | 0.613 (−6.8%) |
| TabBar | 0.795 (−7.2%) | 0.504 (−14.2%) | 0.616 (−11.5%) | 0.540 (−10.1%) |
| Table | 0.740 (−6.0%) | 0.762 (−9.1%) | 0.750 (−7.6%) | 0.745 (−6.4%) |
| Radio_Selected | 0.857 (−8.7%) | 0.705 (−14.0%) | 0.773 (−11.6%) | 0.779 (−3.3%) |
| Radio_Unselected | 0.739 (−6.8%) | 0.797 (−12.1%) | 0.766 (−9.3%) | 0.776 (−4.5%) |
| Checkbox_Unchecked | 0.873 (−4.3%) | 0.748 (−9.3%) | 0.805 (−7.0%) | 0.791 (−5.3%) |
| Checkbox_Checked | 0.781 (−8.0%) | 0.508 (−12.5%) | 0.615 (−10.7%) | 0.526 (−7.0%) |
| Tree | 0.678 (−6.9%) | 0.722 (−8.3%) | 0.699 (−7.6%) | 0.707 (−4.2%) |
| Image | 0.884 (−4.7%) | 0.755 (−10.0%) | 0.814 (−7.5%) | 0.745 (−9.1%) |
| Text | 0.870 (−6.2%) | 0.701 (−12.0%) | 0.776 (−9.4%) | 0.740 (−7.7%) |
| Label_of_the_Textarea | 0.846 (−5.5%) | 0.418 (−12.1%) | 0.559 (−10.0%) | 0.431 (−9.4%) |
| Description_List | 0.675 (−3.5%) | 0.719 (−10.2%) | 0.696 (−6.7%) | 0.664 (−5.5%) |
| Legend | 0.761 (−4.8%) | 0.550 (−14.9%) | 0.638 (−10.7%) | 0.610 (−5.7%) |
| Horizontal_Axis | 0.158 (+35.0%) | 0.086 (+60.1%) | 0.111 (+51.3%) | 0.110 (+759.3%) |
| Chart | 0.564 (−6.4%) | 0.676 (−10.4%) | 0.614 (−8.2%) | 0.665 (−6.9%) |
| Graph | 0.579 (−4.6%) | 0.471 (−18.0%) | 0.519 (−12.0%) | 0.517 (−9.6%) |
| Vertical_Axis | 0.736 (−3.1%) | 0.166 (+40.6%) | 0.270 (+32.6%) | 0.213 (+70.4%) |
| Date_area | 0.527 (−5.3%) | 0.503 (−14.4%) | 0.514 (−10.0%) | 0.529 (−6.7%) |

### 4.4. Computational cost

Table 7 presents the number of parameters, average training time for one epoch, and average inference time for one sample, separately for different variations of the multi-modal UI detection model and the baseline. The reported processing times were measured on a server equipped with a NVIDIA Tesla T4 GPU, 16 processing cores, and 110 GB of RAM. One epoch of training processed 14,155 images with a batch size of five. We measured the training time for every epoch and computed the average over all epochs as the average training time for one epoch. Moreover, we ran the model in inference mode on the 2,000 test images (one by one, not in batches), measured the inference time for every image, and computed the average over all images as the average inference time for one image.

The computational cost analysis, summarized in Table 7, highlights the trade-offs introduced by the multi-modal enhancements to the baseline YOLO model. The baseline model has approximately 142.4 million parameters, with a training time of 37.8 minutes per epoch and an inference time of 0.83 seconds per sample. As cross-attention modules are added, the model size, training time, and inference time all increase. For instance, the multi-modal model with element-wise addition and five cross-attention layers has 156 million parameters, requiring 43.2 minutes per epoch for training and 0.98 seconds per sample for inference. The weighted sum fusion strategy introduces only a negligible number of additional parameters compared to element-wise addition, resulting in very similar computational demands. This makes weighted sum an appealing middle-ground, offering improved performance over element-wise addition with minimal extra cost.

**Table 7.** Number of parameters, average training time, and average inference time measured for different variations of the multi-modal UI detection model and the baseline.

| Model | | | Model parameters | Average training time (1 epoch) | Average inference time (1 sample) |
|---|---|---|---|---|---|
| Baseline | | | 142,387,584 | 37.8 min | 0.83 sec |
| Multi-modal | Element-wise addition | 3 cross-attention | 149,725,184 | 41.1 min | 0.89 sec |
| | | 4 cross-attention | 152,102,784 | 42.7 min | 0.93 sec |
| | | 5 cross-attention | 155,976,384 | 43.2 min | 0.98 sec |
| | Weighted sum | 3 cross-attention | 149,729,024 | 41.8 min | 0.90 sec |
| | | 4 cross-attention | 152,107,904 | 43.1 min | 0.93 sec |
| | | 5 cross-attention | 155,983,424 | 43.5 min | 0.98 sec |
| | Convolutional fusion | 3 cross-attention | 152,598,144 | 44.2 min | 0.95 sec |
| | | 4 cross-attention | 155,796,864 | 44.5 min | 0.98 sec |
| | | 5 cross-attention | 161,516,544 | 46.3 min | 1.05 sec |

Moreover, the convolutional fusion strategy is the most computationally expensive, reflecting its increased complexity. The model with five cross-attention layers and convolutional fusion reaches 161.5 million parameters, 46.3 minutes per epoch for training, and 1.05 seconds per sample for inference, representing approximately a 13% increase in parameters and a 26% increase in inference time over the baseline. These increases, while notable, are justified by the consistently superior performance achieved by convolutional fusion. However, they also underscore the need for careful trade-off considerations when deploying the model in latency-sensitive or resource-constrained environments. Ultimately, the choice of fusion strategy should balance performance requirements with computational constraints, with simpler methods like element-wise or weighted sum being preferable for lightweight deployments, and convolutional fusion better suited for performance-critical applications.

## 5. Conclusion

In this work, we proposed a novel multi-modal extension of the YOLOv5 framework for UI control detection. Our model addresses the inherent challenges of UI element detection, such as visual ambiguity, design variability, and lack of contextual information, by integrating GPT-generated textual descriptions with visual features through cross-attention modules. This design allows semantic cues from text to dynamically guide the detection process, improving the model's ability to distinguish between visually similar or semantically complex UI controls.

We validated our approach through extensive experiments on a large-scale dataset of more than 16,000 annotated screenshots spanning 23 UI control classes. Using three different fusion strategies, i.e. element-wise addition, weighted sum, and convolutional fusion, we systematically compared model performance against a strong YOLOv5 baseline. Our results consistently showed that multi-modal integration outperforms the vision-only model, with convolutional fusion achieving the best overall performance. Notably, the proposed model delivered significant improvements for difficult classes such as *Horizontal_Axis*, *Vertical_Axis*, *Graph*, *Chart*, *Image*, and *Checkbox_Checked*, which were poorly detected by purely visual methods.

The findings of this study demonstrate that incorporating textual semantics can substantially enhance UI element detection, especially in cases where pixel-based features alone are insufficient. Beyond advancing detection accuracy, our multi-modal framework has practical implications for automated software testing, accessibility tools, and UI analytics. By offering more reliable and context-aware detection, the proposed approach can enable more robust testing pipelines, improve screen-reader technologies for users with disabilities, and provide deeper insights into user interactions with complex software systems.

At the same time, our work highlights several challenges and opportunities for future research. The sensitivity of the model to inaccurate or incomplete textual input, as revealed by ablation studies, underscores the need for developing mechanisms to handle noisy language descriptions. Moreover, while convolutional fusion provides the best performance, its computational overhead may pose deployment challenges in resource-constrained environments. Future directions include exploring more efficient fusion mechanisms, robustness training against noisy text, leveraging larger and more diverse datasets, and extending our approach to interactive or dynamic UIs. These lines of investigation hold promise for further strengthening multi-modal UI control detection and broadening its applicability across real-world scenarios.

**Appendix A**

The prompt given to the GPT model for extracting textual descriptions of screenshot images comes in the following. We used this prompt in all the experiments reported in the paper.

```
Prompt:

Find the user interface controls as many as you can and denote their label,
size, relative position, shape, and color accordingly. Present the results as a
list, and no more other explanations, such as:

*Icon*
  – Label: "VISION" logo
  – Size: Approximately 100x30 pixels
  – Position: Top left corner of the main content area.
  – Shape: rectangle, square, rounded, narrow, wide, ...
  – Color: white, black, ...

*Window*
  – Label: "Financial Reporting Center"
  – Size: Full width of the main content area
  – Position: Below the "VISION" logo
  – Shape: rectangle, square, rounded, narrow, wide, ...
  – Color: white, black, ...

*Icon*
  – Label: Various icons in the toolbar (e.g., home, star, bell, user, search, help)
  – Size: Approximately 20x20 pixels each
  – Position: Top right corner of the main content area
  – Shape: rectangle, square, rounded, narrow, wide, ...
  – Color: white, black, ...

*Dropdown*
  – Label: User dropdown ("Max Mustermann")
  – Size: Approximately 100x20 pixels
  – Position: Top right corner of the main content area, next to the icons
  – Shape: rectangle, square, rounded, narrow, wide, ...
  – Color: white, black, ...

*Scrollbar*
  – Label: Vertical scrollbar for the table
  – Size: Full height of the table
  – Position: Right side of the table
  – Shape: rectangle, square, rounded, narrow, wide, ...
```

```
  - Color: white, black, ...
The user interface control classes are Icon, Dropdown, Button, Menu, Input,
List, TabBar, Table, Radio_Selected, Radio_Unselected, Checkbox_Unchecked,
Checkbox_Checked, Tree, Image, Text, Label_of_the_Textarea, Description_List,
Legend, Horizontal_Axis, Chart, Graph, Vertical_Axis, and Date_area.
```

As an example of textual descriptions generated by the GPT model, the following description was generated for the software screenshot shown by Figure 7.

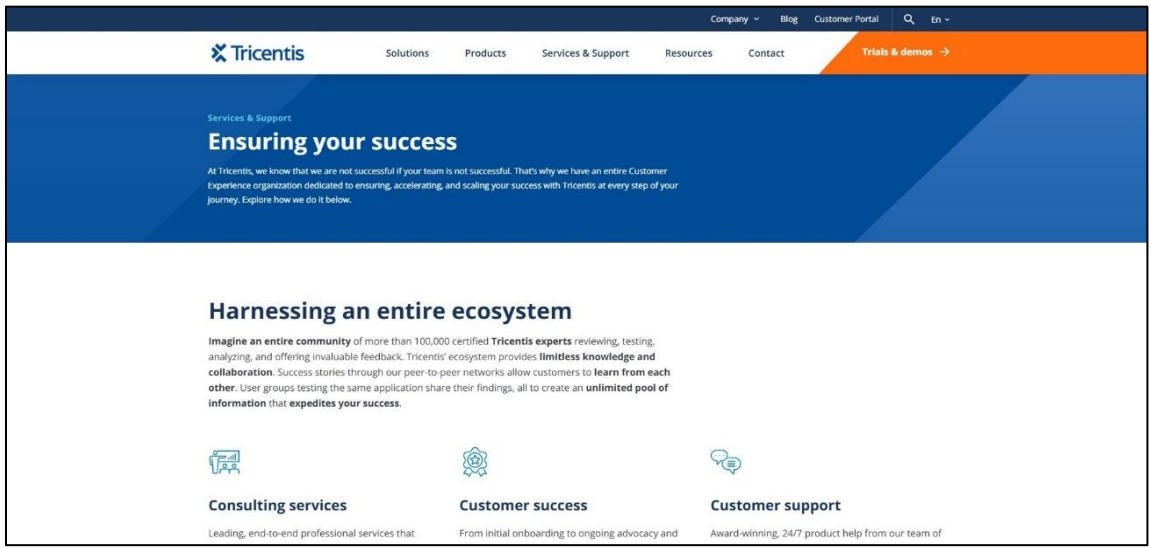

**Figure 7.** A sample screenshot given to the GPT model for generating a textual description of the image.

```
Textual description by GPT:
Icon
Label: "Tricentis" logo
Size: ~160x40 px
Position: Top-left corner of the header bar
Shape: wide rectangle
Color: white on dark blue

Menu
Label: Top navigation ("Solutions | Products | Services & Support | Resources
| Contact")
Size: ~720x40 px total, each item ~110x40 px
Position: Top-center of header bar
Shape: horizontal text buttons
Color: dark blue text, white background
```

*Dropdown*

*Label: "Company ▾"*

*Size: ~90x40 px*

*Position: Top-center-left of header bar, left of "Blog"*

*Shape: rectangle*

*Color: white text, dark blue background*

*Text*

*Label: "Blog"*

*Size: ~60x40 px*

*Position: Top-right of center nav, left of "Customer Portal"*

*Shape: rectangle*

*Color: white text, dark blue background*

*Text*

*Label: "Customer Portal"*

*Size: ~140x40 px*

*Position: Top-right area, left of search icon*

*Shape: rectangle*

*Color: white text, dark blue background*

*Icon*

*Label: Search (magnifying glass)*

*Size: ~24x24 px*

*Position: Top-right header bar, left of language selector*

*Shape: square*

*Color: white*

*Dropdown*

*Label: Language ("En ▾")*

*Size: ~60x40 px*

*Position: Top-right header bar*

*Shape: rectangle*

*Color: white text, dark blue background*

*Button*

*Label: "Trials & demos →" (CTA ribbon)*

*Size: ~260x54 px visible*

*Position: Top-right overlay ribbon across header*

*Shape: wide trapezoid/angled rectangle*

*Color: orange with white text*

*Text*

*Label: "Services & Support" (section label)*

*Size: ~180x20 px*

*Position: Upper hero band, left side*

*Shape: text line*

*Color: light blue text*

*Text*

*Label: "Ensuring your success" (hero H1)*

*Size: ~520x60 px*

*Position: Hero area, left side below label*

*Shape: multiline text*

*Color: white*

*Text*

*Label: Hero description paragraph*

*Size: ~860x60 px*

*Position: Hero area, left side under H1*

*Shape: paragraph block*

*Color: light gray/white*

*Image*

*Label: Blue diagonal hero background*

*Size: Full width x ~280 px height*

*Position: Top of main content under header*

*Shape: rectangle with diagonal gradient*

*Color: dark-to-light blue*

*Text*

*Label: "Harnessing an entire ecosystem" (H2)*

*Size: ~640x50 px*

*Position: Main content, left aligned under hero*

*Shape: text line*

*Color: dark navy*

*Text*

*Label: Intro paragraph under H2*

*Size: ~980x120 px*

*Position: Main content, left aligned under H2*

*Shape: paragraph block*

*Color: dark gray*

*Icon*

*Label: Consulting services pictogram*

*Size: ~44x44 px*

*Position: Left column icon row, under intro*

*Shape: rounded pictogram*

*Color: light blue*

*Text*

*Label: "Consulting services" (H3)*

*Size: ~240x28 px*

*Position: Left column title under its icon*

*Shape: text line*

*Color: dark navy*

*Text*

*Label: Consulting services blurb*

*Size: ~360x80 px*

*Position: Left column body under title*

*Shape: paragraph block*

*Color: dark gray*

*Icon*

*Label: Customer success rosette icon*

*Size: ~44x44 px*

*Position: Middle column icon row, aligned with left icon*

*Shape: rounded pictogram*

*Color: light blue*

*Text*

*Label: "Customer success" (H3)*

*Size: ~240x28 px*

*Position: Middle column title under its icon*

*Shape: text line*

*Color: dark navy*

*Text*

*Label: Customer success blurb*

*Size: ~360x80 px*

*Position: Middle column body under title*

*Shape: paragraph block*

*Color: dark gray*

*Icon*

*Label: Customer support chat bubbles icon*

*Size: ~44x44 px*

*Position: Right column icon row, aligned with others*

*Shape: rounded pictogram*

*Color: light blue*

*Text*

*Label: "Customer support" (H3)*

*Size: ~240x28 px*

*Position: Right column title under its icon*

*Shape: text line*

*Color: dark navy*

*Text*

*Label: Customer support blurb*

*Size: ~360x80 px*

*Position: Right column body under title*

*Shape: paragraph block*

*Color: dark gray*

*Icon*

```
Label: Caret indicators next to dropdown labels ("Company", "En")

Size: ~10x10 px each

Position: Adjacent to respective labels

Shape: triangle

Color: white

Text

Label: "At Tricentis, we know that we are not successful…" (secondary hero copy)

Size: ~980x70 px

Position: Hero area, left below H1

Shape: paragraph block

Color: light gray

Menu

Label: Primary nav grouping (Solutions, Products, Services & Support, Resources, Contact)

Size: ~720x40 px

Position: Same line as "Tricentis" in the header

Shape: horizontal tabs

Color: white text; dark blue background

Description_List

Label: Three-column offerings (Consulting services, Customer success, Customer support)

Size: Full width ~1180x260 px

Position: Below intro paragraph

Shape: three-column grid

Color: dark navy headings, dark gray text, light blue icons
```

Figure 8 shows the UI controls detected within the sample screenshot by the single-modal YOLO model. Figure 9 shows the UI controls detected within the sample screenshot by the multi-modal YOLO model fine-tuned on images and textual descriptions. As can be seen, the multi-modal detection model has been able to detect more UI controls, and also more accurately than the single-modal model.

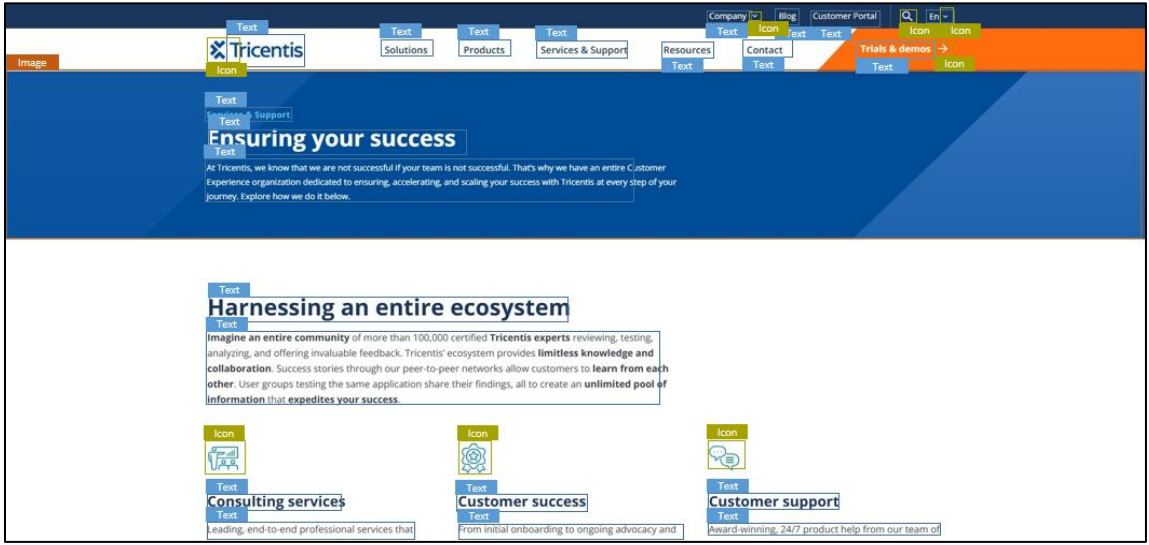

**Figure 8.** The UI controls detected within the sample screenshot by the single-modal YOLO model.

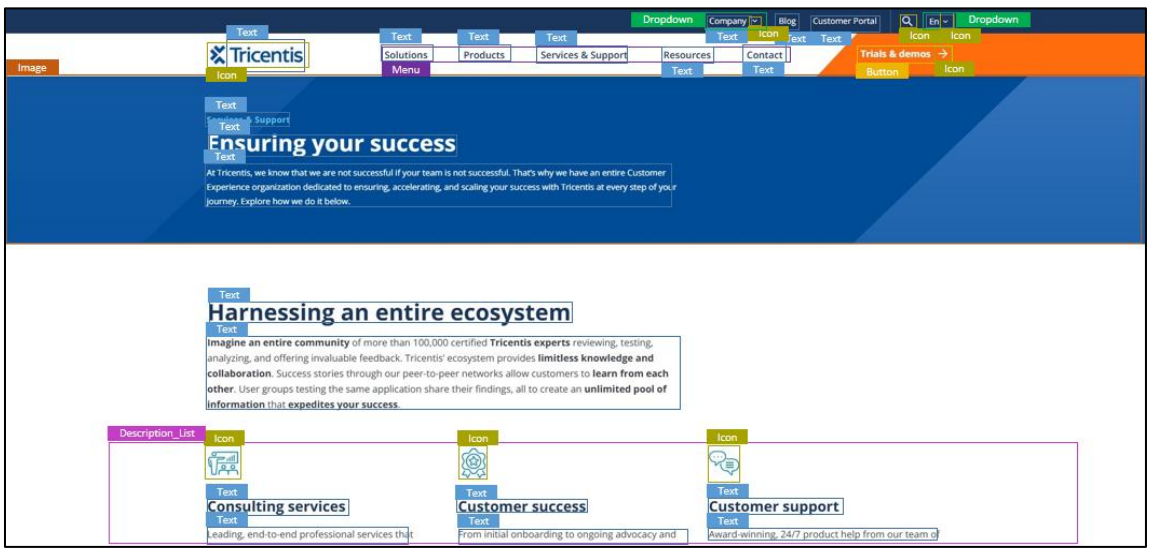

**Figure 9.** The UI controls detected within the sample screenshot by the multi-modal YOLO model fine-tuned on images and textual descriptions.

37